\def\year{2021}\relax
\documentclass[letterpaper]{article} 
\usepackage{aaai21}  
\usepackage{times}  
\usepackage{helvet} 
\usepackage{courier}  
\usepackage[hyphens]{url}  
\usepackage{graphicx} 
\usepackage{natbib}  
\usepackage{caption} 
\usepackage{comment}
\excludecomment{omitext}

\usepackage{times,colordvi,bbm}
\usepackage{tikz}
\usetikzlibrary{patterns}
\usepackage{subfig}
\usepackage[latin1]{inputenc}
\usepackage[fleqn]{amsmath}
\usepackage{amsfonts,amssymb,amsbsy,amscd,amsthm}
\usepackage[mathscr]{eucal}
\usepackage{graphicx}
\usepackage{booktabs}
\usepackage{tabularx,multirow}
    \newcolumntype{L}{>{\raggedright\arraybackslash}X}
    \newcolumntype{C}{>{\centering\arraybackslash}X}
\usepackage{mathrsfs}
\usepackage{array}
\usepackage{verbatim}
\usepackage{enumitem}
\usepackage[linesnumbered,ruled,lined,boxed]{algorithm2e}
\RestyleAlgo{boxruled}
\SetKwRepeat{Do}{do}{while} 

\usepackage{listings}
\usepackage[off]{auto-pst-pdf}
\usepackage{ifthen}
\usepackage{mathtools}
\DeclarePairedDelimiter{\nm}{\lVert}{\rVert}

\mathchardef\breakingcomma\mathcode`\,
{\catcode`,=\active
  \gdef,{\breakingcomma\discretionary{}{}{}}
}

\renewcommand{\vec}[1]{{\ensuremath{\boldsymbol{\mathrm #1}}}}
\newcommand{\vdot}{{\ifnum\thedotStyle=0{\ensuremath\cdot}\else
    {\boldsymbol{\mathsf{\ensuremath\cdot}}}\fi}}

\newcommand{\bu}{\mathbf{u}}
\newcommand{\bv}{\mathbf{v}}

\newcommand{\bz}{\mathbf{z}}

\newcommand{\bo}{\boldsymbol{\omega}}

\newcommand{\grad}{\nabla}
\newcommand{\pd}[2]{\frac{\partial{#1}}{\partial{#2}}}
\newcommand{\pdt}[1]{\pd{#1}t}
\newcommand{\ddt}[1]{\frac{d \,{#1}}{dt}}

\newcommand{\R}{\ensuremath{\mathbb{R}}}
\newcommand{\RR}{\ensuremath{\mathcal{R}}}

\newcommand{\lrp}[1]{\left( #1 \right)}
\newcommand{\lb}{\left[}
\newcommand{\rb}{\right]}

\newcommand{\lB}{\left\{}
\newcommand{\rB}{\right\}}

\newcommand{\ie}{\textit{i}.\textit{e}.\ }

\newcommand{\strongRes}{\boldsymbol{\mathfrak{R}}}

\theoremstyle{remark}

\graphicspath{{./figures/}}

\title{Neural Ordinary Differential Equations for Data-Driven Reduced Order Modeling of Environmental Hydrodynamics}
\author{
    Sourav Dutta,\textsuperscript{\rm 1}
    Peter Rivera-Casillas,\textsuperscript{\rm 2}
    Matthew W. Farthing,\textsuperscript{\rm 1}
    \\
}
\affiliations{
    \textsuperscript{\rm 1} U.S. Army Engineer Research and Development Center, Coastal and Hydraulics Laboratory, Vicksburg, MS, \\
    \textsuperscript{\rm 2} U.S. Army Engineer Research and Development Center, Information and Technology Laboratory, Vicksburg, MS, \\
    \textsuperscript{\rm 1, \rm 2} \{sourav.dutta, peter.g.rivera-casillas, matthew.w.farthing\}@erdc.dren.mil

}

\begin{document}

\maketitle

\begin{abstract}
Model reduction for fluid flow simulation continues to be of great
interest across a number of scientific and engineering
fields. Here, we explore the use of Neural Ordinary Differential Equations, a recently 
introduced family of continuous-depth, differentiable networks 
\cite{Chen2018}, as a way to propagate latent-space dynamics in reduced order
models. We compare their behavior with two classical
non-intrusive methods based on proper orthogonal decomposition and
radial basis function interpolation as well as dynamic mode decomposition. The
test problems we consider include incompressible flow around a
cylinder as well as real-world applications of shallow water
hydrodynamics in riverine and estuarine systems. Our findings 
indicate that Neural ODEs provide an elegant framework for stable 
and accurate evolution of latent-space dynamics with a promising 
potential of extrapolatory predictions. However, in order to 
facilitate their widespread adoption for large-scale systems, 
significant effort needs to be directed at accelerating their training times. This 
will enable a more comprehensive exploration of the hyperparameter space for 
building generalizable Neural ODE approximations over a wide range of system dynamics.
\end{abstract}

\section{Introduction}
Despite the trend of hardware improvements and significant gains in the algorithmic efficiency of standard discretization procedures, \textit{high-fidelity} numerical simulation of engineering systems governed by nonlinear partial differential equations still pose a prohibitive computational challenge \cite{Quarteroni_Manzoni_etal_16} for several decision-making applications involving control \cite{PBK2016}, optimal design and multi-fidelity optimization \cite{PWG2016}, and/or uncertainty quantification \cite{SM2013}. \textit{Reduced order models} (ROMs) offer a valuable alternative way to simulate such dynamical systems with considerably reduced computational cost \cite{Benner_Gugercin_etal_15}.

\textit{Reduced basis} (RB) methods
\cite{Quarteroni_Manzoni_etal_16} constitute a family of widely popular ROM techniques that are usually implemented with an offline-online decomposition paradigm. The \textit{offline} stage involves the construction of a solution-dependent, linear basis space spanned by a set of RB ``modes'', which are extracted from a collection of high-fidelity solutions, also called \textit{snapshots}. The RB ``modes'' can be thought of as a set of global basis functions spanning a linear subspace that can be used to approximate the dynamics of the high-fidelity model. The most well known method to extract the reduced basis is called \textit{proper orthogonal decomposition} (POD) \cite{Sirovich_87,BHL1993}, which is particularly effective when the coherent structures of the flow can be hierarchically ranked in terms of their energy content. 

In the \textit{online} stage of traditional RB methods, a linear combination of the reduced order RB modes is used to approximate the high-fidelity solution for a new configuration of flow parameters. The procedure adopted to compute the expansion coefficients leads to the classification of these methods into two broad categories: \textit{intrusive} and \textit{non-intrusive}. In an intrusive RB method, the expansion coefficients are determined by the solution of a reduced order system of equations, which is typically obtained via a Galerkin or Petrov-Galerkin projection of the high-fidelity (full-order) system onto the RB space \cite{LFK2017}. Typically this projection and solution involves modification of high-fidelity simulators and hence the label intrusive. For linear systems, Galerkin projection is the most popular choice. However, in the presence of nonlinearities, an affine expansion of the nonlinear (or non-affine) differential operator must be recovered in order to make the evaluation of the projection-based reduced model independent of the number of DOFs of the high-fidelity solution. 

Several different techniques, collectively referred to as hyper-reduction methods \cite{AZCF2015}, have been proposed to address this problem. These include the empirical interpolation method (EIM), its discrete counterpart DEIM \cite{CS2010}, ``gappy POD" \cite{W2006}, as well as the residual DEIM method \cite{XFBPNDH2014}. Beyond the need for hyper-reduction to recover efficiency, in complex nonlinear problems it is also common that some of the intrinsic structures present in the high-fidelity model may be lost during order reduction using Galerkin projection-based approaches. This is because the Galerkin projection approach inherently assumes that the residual generated by the truncated representation of the high-fidelity model is orthogonal to the reduced basis space which leads to the loss of higher order nonlinear interaction terms in the reduced representation. This can result in qualitatively wrong solutions or instability issues \cite{AF2012}. As a remedy, Petrov-Galerkin projection based approaches have been proposed {\cite{CBF2011,CFCA2013,FPNEDX2013}}.  

An alternative family of methods to address the issues of instability and loss of efficiency in the intrusive ROM frameworks is represented by \textit{non-intrusive} reduced order models (NIROMs), and forms the subject of this study. The primary advantage of this class of methods is that complex modifications to the source code describing the physical model can be avoided, thus
making it easier to develop reduced models when the legacy or proprietary source codes are not available. In these methods, instead of a Galerkin-type projection, the expansion coefficients for the reduced solution are obtained via interpolation on the space of a reduced basis extracted from snapshot data. However, since the reduced dynamics generally belong to nonlinear, matrix manifolds, a variety of interpolation techniques have been proposed that are capable of enforcing the constraints characterizing those manifolds. Regression-based non-intrusive methods have been proposed that, among others, use artificial neural networks (ANNs), in particular multi-layer perceptrons \cite{HU2018}, Gaussian process regression (GPR) \cite{GH2019}, and radial basis function (RBF) \cite{ADN2013} to perform the interpolation. 

Here, we will explore an alternative approach to propagating latent-space dynamics based on Neural ODEs, which are a family of continuous-depth, differentiable networks that can be seen as an extension of ResNets in the limit of a zero discretization step size \cite{Dupont2019}. Details of our approach follow below. In addition, we consider two NIROM techniques - a) based on linear dimension reduction via POD and latent space evolution via Radial Basis Functions (RBF), and b) Dynamic Mode Decomposition (DMD) that will serve as the benchmarks in our numerical experiments to provide comparisons with the Neural ODE approach. We then proceed with several numerical experiments based on incompressible flow around a cylinder and shallow water hydrodynamics in order to evaluate the methods' performance for fast replay applications in complex fluid-dynamics problems.

\section{Methodology}
The standard ROM development framework usually consists of three stages:
\begin{enumerate}
    \item identification of a low-dimensional latent (or reduced-order) space,
    \item determining a latent-space representation of the nonlinear dynamical system in terms of the reduced basis and modeling the evolution of the system of modal coefficients, and
    \item reconstruction in the high-fidelity space for validation and analysis.
\end{enumerate}
Machine learning techniques can be introduced at any of these stages. For example, many works have explored the use of deep learning-based approaches like autoencoders as a way to introduce a nonlinear alternative for dimension reduction \cite{Lusch2018, Ghorbanidehno2021}. Combining these methods with data-driven latent-space propagation (for example via fully connected or recurrent neural networks) leads to a fully non-intrusive approach \cite{Gonzalez_Balajewicz_2018}. On the other hand, one can also combine nonlinear dimension reduction with intrusive projection to create a hybrid method \cite{Lee2020,Kim2020}. In this work, we study three different data-driven strategies for accurate learning of system dynamics within the context of linear dimension reduction. In the first two methods we adopt the POD technique for identification of an optimal global basis. For the latent space evolution, we utilize a kernel-based multivariate interpolation method called radial basis function (RBF) interpolation, and a machine learning strategy designed for sequential learning of time-series data called neural ordinary differential equations (NODE). In the third strategy, the three stages of ROM development are combined together by using a classical modal decomposition technique called the dynamic mode decomposition (DMD), that is supported by rigorous mathematical analysis of Koopman mode theory.

\subsection{Proper orthogonal decomposition}
POD is a popular technique for dimension reduction \cite{AS2001} of the solution manifold of a dynamical system by determining a linear reduced space spanned by an orthogonal basis with an associated energetic hierarchy. \cite{Taira2020} provides an excellent overview of POD as well as a comparison with other dimension-reduction techniques. 

Consider a \textit{snapshot matrix}
$\vec{S} = [\widehat{\bv}^1, \ldots, \widehat{\bv}^M] \in \R^{N\times M}$ containing a collection of $M$ high-fidelity snapshots of the solution manifold from time $t=0$ to $t=T$ such that $\widehat{\bv}^k \in \R^N$ is the $k^{th}$ snapshot with the temporal mean value removed, \ie, $\widehat{\bv}^k = \bv^k - \bar{\bv}$ where $\bar{\bv} = \sum_{i=1}^M\frac{\bv^i}{M}$ is the time-averaged solution. The goal of the POD procedure is to identify a linear subspace $\chi = \text{span} \lB \psi^1, \ldots, \psi^r \rB, \; (r \ll M)$ which approximates the solution manifold optimally with respect to the $L^2$-norm.

The POD bases can be efficiently extracted by performing a ``thin" singular value decomposition (SVD) of the snapshot matrix $\vec S = \widetilde{\vec \Theta} \widetilde{\boldsymbol{\Sigma}} \widetilde{\vec \Psi}^T$,
where $\widetilde{\boldsymbol{\Sigma}} = \text{diag}(\sigma_1,\ldots,\sigma_\RR)$ is a $\RR\times \RR$ diagonal matrix containing the singular values arranged in decreasing order of magnitude, $\sigma_1 \geq \sigma_2 \ldots \geq \sigma_\RR$ and $\RR < \min \{N,M\}$ is the rank of $\vec S$. $\widetilde{\vec \Theta}$ and $\widetilde{\vec \Psi}$ are $N\times \RR$ and $M \times \RR$ matrices respectively, whose columns are the orthonormal left and right singular vectors of $\vec S$ such that $\widetilde{\vec \Theta}^T \widetilde{\vec \Theta} = \vec I_{\RR} = \widetilde{\vec \Psi}^T \widetilde{\vec \Psi}$.
The columns $\vec \theta_n$ of the matrix $\widetilde{\vec \Theta}$ are ordered corresponding to the singular values $\sigma_n$ and these provide the desired POD basis. 
Let ${\vec \Theta}$ denote the matrix of the first $m$ columns of $\widetilde{\vec \Theta}$, $\vec \Psi$ be the matrix containing the first $m$ rows of $\widetilde{\vec \Psi}$, and ${\boldsymbol{\Sigma}}$ be a diagonal matrix containing the first $m$ singular values from $\widetilde{\boldsymbol{\Sigma}}$, then the high-fidelity solution $\bv^n$ at time $t^n$ can be approximated as,
\begin{align}\label{eq:pod-basis}
    \bv^n \approx \bar{\bv} + {\vec \Theta} \bz^n = \bar{\bv} + \sum_{i=1}^m z_i^n \vec \theta_i,
\end{align}
where $\bz^n \in \R^m$ is a vector of modal coefficients with respect to the reduced basis. The modal coefficient matrix $\vec Z = \vec \Theta^T \vec S $ constitutes our training data for the latent space learning methods. Due to the Eckart-Young-Mirsky theorem, the POD basis provides an optimal rank-$m$ approximation $\widehat{\vec S} = {\vec \Theta} {\boldsymbol{\Sigma}} {\vec \Psi}^T$ of the snapshot matrix $\vec S$ with a desired level of accuracy, $\tau_{POD}$.

The POD method has been successfully applied in statistics \cite{J1986}, signal analysis and pattern recognition \cite{DM2008}, ocean models \cite{VH2006}, air pollution models \cite{FZPPBN2014}, convective Boussinesq flows \cite{SB2015}, and Shallow Water Equation (SWE) models {\cite{SSN2014,LFKG2016}}.

\subsection{Latent space evolution}
In this section, we outline two non-intrusive methods for modeling the evolution of time-series data in the latent space defined by the POD basis. RBF interpolation is a classical, data-driven, kernel-based method for computing an approximate continuous response surface that aligns with the given multivariate data. The second technique called NODE is a neural-network based method to predict the continuous evolution of a vector $\vec c$ over time, that is designed to preserve memory effects within the architecture. 

\subsubsection{Radial basis function interpolation}
For simplicity, let the time evolution of the modal coefficients $\bz$ be represented as a semi-discrete dynamical system,
\begin{align}\label{eq:POD-semi-discrete}
\dot{\bz} = \vec f(\bz,t), \; \text{ with } \vec z^0 = \vec {\Theta}^T\lrp{\vec v^0 - \bar{\vec v}}
\end{align}
where all the information about the temporal dynamics including the effects of any numerical stabilization of the high-fidelity solver and all the nonlinear terms are embedded in $\vec f(\bz,t)$. 
In the POD-RBF NIROM framework \cite{DFPSP2020}, instead of the Galerkin projection, the components of the time derivative function $f_j (j=1,\ldots,m)$ are approximated using RBF interpolation. 

Let $F_j$ denote a RBF approximation of the time derivative function $f_j$, which is defined by a linear combination of $N_i$ instances of a radial basis function $\phi$,
\begin{equation}\label{eq:nirom-rbf-interpolant}
F_j (\vec z) = \sum_{k=1}^{N_i} \alpha_{j,k} \, \phi \lrp{\nm{\vec z - \widehat{\vec z}_k}}, \quad j=1,\ldots,m,
\end{equation}
where $\{ \widehat{\vec z}_k \, | \, k = 1,\ldots, N_i \}$ denotes the set of interpolation centers and $\alpha_{j,k}$ $(k = 1,\ldots, N_i)$ is the unknown interpolation coefficient corresponding to the $k^{th}$ center for the $j^{th}$ component of the modal coefficient.
These interpolation coefficients are computed by enforcing the interpolation function $F_j$ to exactly match the time derivative of the modal coefficients at $N_e$ test points ($N_e \geq N_i$).
Choosing the centers and the test points identically from the set of snapshot modal coefficients as $\{\vec z^l \,|\, l = 0, \ldots, M-1\}$ such that $N_i = N_e = M$, and making some simplifying assumptions leads to a symmetric, linear system of $M$ equations to solve for the unknown interpolation coefficients, $\alpha_{j,k}$  
\begin{gather}\label{eq:interp-cond}
 \vec A \vec \alpha^j=\boldsymbol{g}^j, \; \text{ for } j=1,\ldots,m,
\end{gather}
where 
\begin{align*}
    [A_{n,k}] = [\phi\lrp{\nm{\vec z^n - \vec z^k}}],  \quad  n,k = 0, \ldots M-1,\\
    \vec \alpha^j = [\alpha_{j,0}, \ldots, \alpha_{j,M-1}]^T , \;  
    \boldsymbol{g}^j = [g_{j,0}, \ldots, g_{j,M-1}]^T.
\end{align*}
The coefficients $\vec \alpha^j$ define a unique RBF interpolant which can then be used to approximate eq.~(\ref{eq:POD-semi-discrete}) and generate a non-intrusive model for the evolution of the modal coefficients 
In this work, a first-order forward Euler scheme has been employed for the discretization of the time derivative, and a strictly positive-definite Mat\'{e}rn $C^0$ kernel, given by $\phi(r) = e^{-cr}$ has been adopted, where $r$ is the Euclidean distance and $c$ is the RBF shape factor \cite{F2007}.

Adopting RBF interpolation for modeling the latent space evolution of the modal coefficients has been shown to be quite successful for nonlinear, time-dependent partial differential equations (PDEs) \cite{XFPH2015,DFPSP2020}, nonlinear, parametrized PDEs \cite{ADN2013,XFPN2017}, and aerodynamic shape
optimization \cite{IQ2013}, to name a few. 

\subsubsection{Neural ordinary differential equations}
Recurrent neural network (RNN) architectures like LSTM and GRU are often employed to encode time-series data and forecast future states, as their internal memory preserving architecture allows them to incorporate state information over a sequence of input data. Although RNNs have seen great success in natural language processing tasks, they have had relatively limited success in high-fidelity scientific computing applications \cite{Ferrandis2019,Wang2020}, as it has been observed that a sequence generated by an RNN may fail to preserve temporal regularity of the underlying signal, and thus may not represent true continuous dynamics. \cite{Chen2018}. With deep neural networks (DNN) such as ResNet, the evolution of the features over the network depth is equivalent to solving an ordinary differential equation (ODE) such as $\frac{dz}{dt} = F(z,\theta)$ using the forward Euler method, and this connection between ResNet's architecture and numerical integrators has been explored in details by \cite{Ruthotto2019} and others. Several other deep learning methods have been proposed for learning ODEs and PDEs. These include using PDE-based network \cite{Long2019}, training DNNs using physics-informed soft penalty constraints \cite{Raissi2019}, and using sparse regularizers and regression \cite{Brunton2016,Champion2019}, to name a few. 

\citet{Chen2018} proposed a 'continuous-depth' neural network called ODE-Net that effectively replaces the layers in ResNet-like architectures with a trainable ODE solver. The memory efficiency and stability of this neural ordinary differential equation (NODE) approach was further improved in \cite{Gholami2019,Dupont2019} and others.
\cite{Maulik2020} applied the NODE framework to obtain latent space closure models for ROMs of a one-dimensional advecting shock problem and a one-dimensional Burgers' turbulence problem that exhibits multiscale behavior in the wavenumber space. Some other notable recent applications of NODE include the identification of ODE or PDE models from time-dependent data \cite{Sun2020}, modeling of irregularly spaced time series data \cite{RCD2019}, modeling of spatio-temporal information in video signals \cite{KVKP2019}. \citet{Finlay2020} used a combination of optimal transport theory and stability regularizations to propose a neural-ODE generative model that can be efficiently trained on large-scale datasets. Here we further explore the application of the POD-NODE methodology to complex, real-world flows characterized by systems of two-dimensional, nonlinear PDEs.

We assume that the time evolution of the modal coefficients of the high-fidelity dynamical system in the latent space can be modeled using a (first-order) ODE, 
\begin{align}\label{node-ivp}
\ddt{\bz} = \mathcal{F}(t, \bz(t)), \text{ with } \bz(0) = \bz^0, \; \bz \in \R^d, d \geq 1.
\end{align}
The goal is to obtain a NN approximation $\widehat{\mathcal{F}}$ of the dynamics function $\mathcal{F}$ such that $\ddt{\bz} \approx \text{net} (t,\bz) = \widehat{\mathcal{F}}(t, \bz, \bo)$. The full procedure can be outlined as follows:
\begin{enumerate}
    \item Compute the time series of modal coefficients $[\bz^0, \ldots, \bz^{M-1}]$ for $t \in \{0,\ldots,M-1\}$ where $\bz^k \in \R^m$.
    \item Initialize a NN approximation for the dynamics function $\widehat{\mathcal{F}}(t, \bz, \bo)$ where $\bo$ represents the initial NN parameters.
    \item The NN parameters are optimized iteratively through the following steps.
    \begin{enumerate}
        \item Compute the approximate forward time trajectory of the modal coefficients by solving eq. (\ref{node-ivp}) using a standard ODE integrator as,
    \begin{align}
        \hat{\bz}^{M-1} = ODESolve(\widehat{\mathcal{F}}, \bo, \bz^0, t^0, t^{M-1})
    \end{align}
    \item The free parameters of the NODE model are $\{\bo,t^0, t^{M-1}\}$. Evaluate the differentiable loss function 
    $\mathcal{L}\left(ODESolve(\widehat{\mathcal{F}}, \bo, \bz^0, t^0, t^{M-1}) - \bz^{M-1}\right)$.
    
    \item To optimise the loss, compute gradients with respect to the free parameters. Similar to the usual backpropagation algorithm, this can be achieved by first computing the gradient $\partial \mathcal{L}/\partial \widehat{\bz}(t)$, and then a performing a reverse traversal through the intermediate states of the ODE integrator. For a memory-efficient implementation, the adjoint method \cite{Chen2018} can be used to backpropagate the errors by solving an adjoint system for the augmented state vector $\vec b = [\pd{\mathcal{L}}{\widehat{\bz}}, \pd{\mathcal{L}}{\bo}, \pd{\mathcal{L}}{t}]^T $ backwards in time from $t^{M-1}$ to $t^0$. 
    \item The gradient $\pd{\mathcal{L}}{\bo}(t=0)$ computed in the previous step is used to update the parameters $\bo$ by using an optimization algorithm like RMSProp or Adam.
    \end{enumerate}
    \item The trained NODE approximation of the dynamics function can be used to compute predictions for the time trajectory of the modal coefficients. 
\end{enumerate}

In this work, we utilize the TFDiffEq (\url{https://github.com/titu1994/tfdiffeq}) library that runs on the Tensorflow Eager Execution platform to train the NODE models. 
Although a single layer architecture guarantees upper-bounds according to the universal approximation theorem \cite{Barron1993}, deeper networks with up to four layers as well as several linear and nonlinear activation functions are also explored due to their possibly improved expressibility for more complex nonlinear dynamics \cite{Zhang_etal_2020}. RMSProp is adopted for loss minimization with an initial learning rate of $0.001$, a staircase decay function with a range of variable decay schedules, and a momentum coefficient of $0.9$. 
NODE predictions of comparable accuracy were obtained for all the numerical experiments by using both the adjoint method as well as by backpropagating gradients directly through the hidden steps of the ode solver. However, for large-scale training data the latter method may lead to memory issues, especially while computing on GPU nodes.

\subsection{Dynamic mode decomposition}
As a final point of comparison, we consider Dynamic mode decomposition (DMD). DMD is a data-driven ROM technique that represents the temporal dynamics of a complex, nonlinear system \cite{Schmid2010,KBBP2016} as the combination of a few linearly evolving, spatially coherent modes that oscillate at a fixed frequency, and which are closely related to the eigenvectors of the infinite-dimensional Koopman operator \cite{K1931,M2013}. Consider the following snapshot matrices containing a few temporally-equispaced snapshots of a high-dimensional dynamical system:
\begin{align*}
    \vec X = \lb \bv^0 \; \bv^1 \; \ldots \bv^{M-1} \rb, \quad \vec X' = \lb \bv^1 \; \bv^2 \; \ldots \bv^{M} \rb
\end{align*}
where $\bv^{k} \in \R^N$ is the $k^{th}$ solution snapshot, $N$ is the spatial degrees of freedom of the discretized system, and $M$ is the total number of temporal snapshots. DMD involves the identification of the best-fit linear operator $\vec A_X$ that relates the above matrices as $\vec X' = \vec A_X \vec X$,
and computing its eigenvalues and eigenvectors. Computing a least-square approximation of $\vec A_X$ using the Moore-Penrose pseudoinverse($^\dagger$) may pose computational challenges due to the size of the discrete dynamical system. For computational efficiency, the exact DMD algorithm (adopted here) avoids computing the Moore-Penrose pseudoinverse($^\dagger$) by projecting the operator on to a reduced space obtained by POD, as outlined in \cite{AK2017}. 

In recent years, Koopman mode theory has provided a rigorous theoretical background for an efficient modal decomposition in problems describing oscillations and other nonlinear dynamics using DMD \cite{Rowley2009}. Several variants of the DMD algorithm have been proposed \cite{PBK2016,KFB2016,ABBN2016,LV2017} and have been successfully applied as efficient ROM techniques for determining the optimal global basis modes for nonlinear, time-dependent problems \cite{DMD,Bistrian2017}. For non-parametrized PDEs, DMD presents an efficient framework that combines all the three stages of ROM development to learn a linear operator in an optimal least square sense. However, this approach cannot be directly applied to parametrized problems \cite{Alsayyari2021}.

\section{Numerical experiments}
In this section, we first assess the performance of different NODE architectures for a benchmark flow problem characterized by the incompressible Navier Stokes equations (NSE), and then further evaluate the relative performance of all three NIROM models for two real-world applications governed by the shallow water equations (SWE). The POD-RBF and DMD NIROM training runs were performed on a Macbook Pro 2018 with a $2.9$ GHz 6-Core Intel Core i9 processor and 32 GB 2400 MHz DDR4 RAM. The NODE models were trained in serial on Vulcanite, a high performance computer at the U.S. Army Engineer Research and Development Center DoD Supercomputing Resource Center (ERDC-DSRC). Vulcanite is equipped with NVIDIA Tesla V100 PCIe GPU accelerator nodes and has 32GBytes memory/node.

\subsection{Flow around a cylinder}
This problem simulates a time-periodic fluid flow through a 2D pipe with a circular obstacle. The flow domain is a rectangular pipe with a circular hole of radius $r=0.05$, denoted by $\Omega = [0,2.2]\times[-0.2, 0.21]\setminus B_r(0.2,0)$. The flow is governed by
\begin{align}
    \pdt{\bu} + \grad \cdot (\bu \otimes \bu) - \nu \Delta \bu + \grad p &= 0, \\
    \grad \cdot \bu &= 0
\end{align}
where $\bu$ denotes the velocity, $p$ the pressure, $\otimes$ is the outer product (dyadic product) given by $\vec a \otimes \vec b = \vec a \vec b^T$, and $\nu=0.001$ is the kinematic viscosity. 
No slip boundary conditions are specified along the lower and upper walls, and on the boundary of the circular obstacle.
A parabolic inflow velocity profile is prescribed on the left wall,
\begin{align}
    \bu(0,y) = \left(4U\frac{(0.21-y)(y-0.2)}{0.41^2},0 \right),
\end{align}
and zero gradient outflow boundary conditions on the right wall. High-fidelity simulation data is obtained with OpenFOAM using an unstructured mesh with $14605$ nodes at $Re=100$, such that the flow exhibits the periodic shedding of von Karman vertices. $313$ training snapshots are collected for $t=[2.5,5.0]$ seconds with $\Delta t=0.008$ seconds, and the NIROM predictions are obtained for $t=[2.5,6.0]$ seconds with $\Delta t=0.002$ seconds.

\begin{table*}[t]
  \centering
  \begin{tabular}{l c c m{1.65cm} m{1.85cm} c c c r}
  \toprule
    Id & Layers & Units & Act.  & LR decay steps, rate & Scaling & Augmented & MSE     & Training\\
    \toprule
    Range & 1-4 & 32-512 & linear, relu, elu, tanh, ... & 5000-25000, 0.1-0.9 &  & & & \\
    \midrule 
    NODE1  & 1      & 256   & elu   & 10000, 0.3  & No      & No        & 8.87e-4 & 24.45 hrs \\
    NODE2  & 1      & 256   & tanh  & 5000,  0.7  & Yes     & No        & 9.01e-4 & 24.56 hrs \\
    NODE3  & 1      & 512   & elu   & 5000, 0.5   & No      & No        & 8.86e-4 & 24.39 hrs \\
    NODE4  & 1      & 256   & tanh  & 10000, 0.25 & Yes     & Yes       & 9.02e-4 & 22.97 hrs \\
    NODE5  & 4      & 64    & tanh  & 5000, 0.5   & Yes     & No        & 9.19e-4 & 27.98 hrs \\
    NODE6  & 1      & 256   & elu   & 10000, 0.1  & No      & No        & 8.87e-4 & 24.13 hrs \\
    NODE7  & 2      & 128   & elu   & 5000, 0.5   & No      & No        & 8.86e-4 & 25.80 hrs \\
    NODE8  & 1      & 512   & tanh  & 5000, 0.5   & Yes     & Yes       & 9.00e-4 & 24.77 hrs \\
    \bottomrule
  \end{tabular}
  \caption{Best NODE architectures for the cylinder example. All models were trained for 50000 epochs using the fourth-order Runge-Kutta solver and the RMSProp optimizer with an initial learning rate of 1e-3 and a momentum of 0.9.}
  \label{tab:node_cylinder}
\end{table*}

\begin{figure}[htb]
    \centering
    \includegraphics[width=0.98\columnwidth]{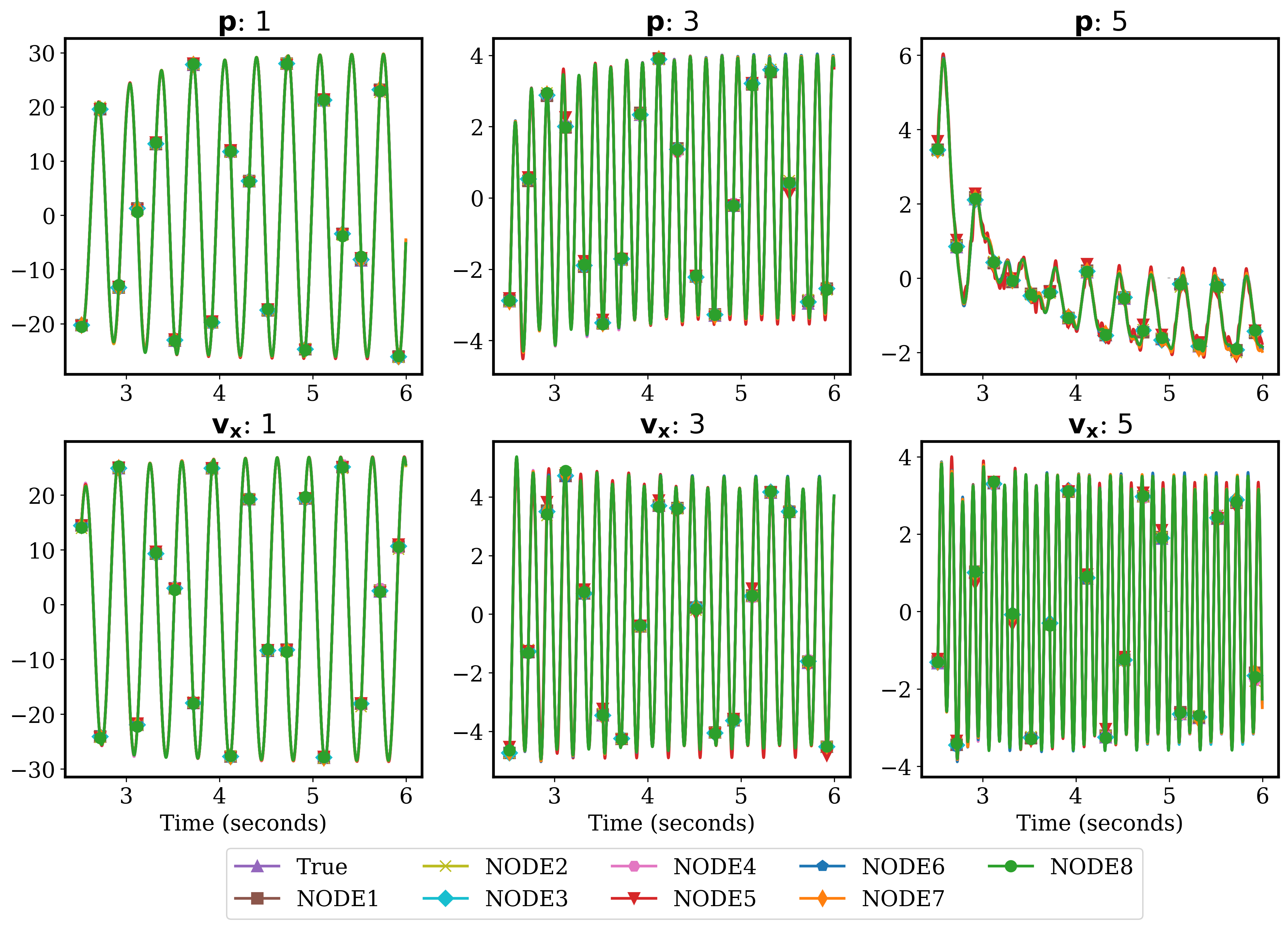}
    \caption{Comparison of NODE models in the latent space for the cylinder example}
    \label{fig:cylinder_modal}
\end{figure}

A large collection of NODE architectures and hyperparameter configurations were trained for 50000 epochs and details of the best 8 models are presented in Table \ref{tab:node_cylinder}. A fourth-order Runge-Kutta solver was found to be the optimal choice in terms of both accuracy and efficiency among all the available solvers ranging from the fixed-step forward Euler and the midpoint solvers to the adaptive-step Dormand-Prince (dopri5) solver. The ``tanh'' and ``elu'' activation functions were found to be the most effective among all the available linear and nonlinear activation functions. Due to the nature of the activation functions, the networks with ``tanh'' activations were found to train better when every element of the input state vector was individually scaled to be bounded in $[-1,1]$, while networks with ``elu'' activations trained better without scaling of input vectors. Augmentation of input states as outlined in \cite{Dupont2019} was found to have no significant impact on the training. The RMSProp optimizer paired with either a step decay function or an exponential decay function were found to be equally effective. However, further numerical experiments are necessary to study the efficiency of alternative first-order and second-order optimization methods. The number of decay steps were varied in discrete increments between 5000 to 25000, and decay rates ranging from 0.1 to 0.9 were studied. It was observed that a lower initial learning rate ($\approx 0.001$) combined with either larger decay steps and smaller decay rates or vice versa led to a desirable training trajectory. Fig. \ref{fig:cylinder_modal} shows the evolution of the $1^{st}, 3^{rd}$, and $5^{th}$ latent-space modal coefficients for the pressure and the x-velocity solutions, obtained using the best 8 NODE models. All the models generate accurate predictions at a finer temporal resolution than the training data, and have excellent agreement with the high-fidelity solution even while extrapolating outside the training data ($5 \leq t \leq 6$ seconds).

\begin{figure}[htb]
  \begin{center}
    \includegraphics[width=0.98\columnwidth]{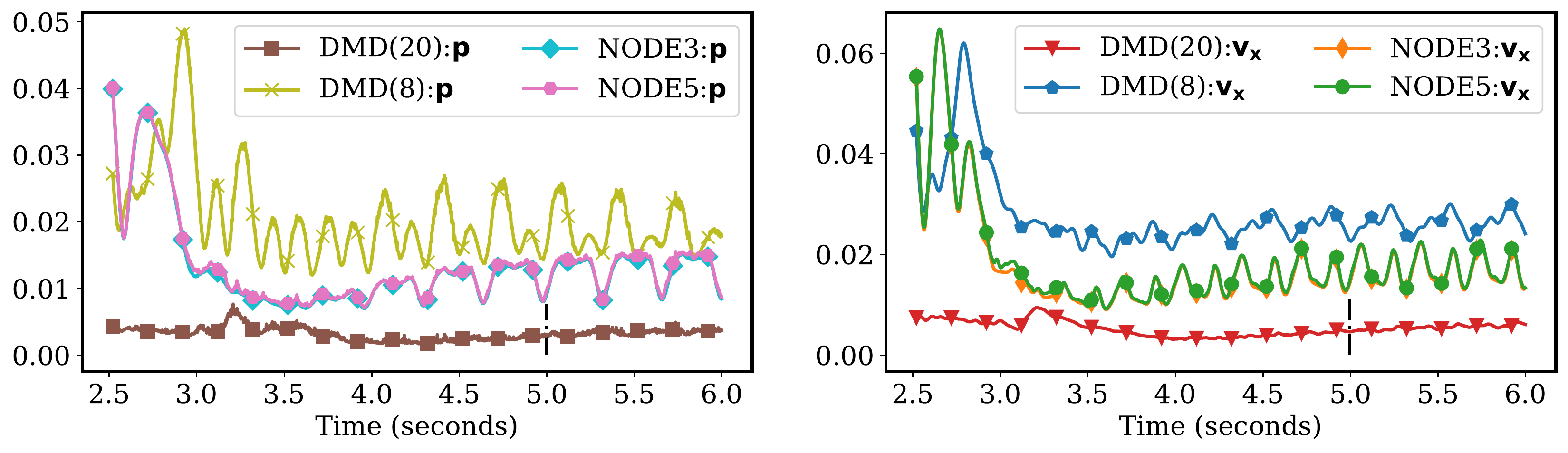}
  \end{center}
  \caption{Comparison of RMSE of best NODE models with DMD for the cylinder example}\label{fig:cylinder_rmse}
\end{figure}

Fig. \ref{fig:cylinder_rmse} compares the time trajectory of the spatial root mean square errors (RMSE) in the high-fidelity space for two of the best NODE models with two DMD NIROM solutions obtained using truncation levels of $r=20$ and $r=8$. It is encouraging to note that even though the NODE solutions are computed using a latent-space representation that is roughly comparable to the DMD solution with a smaller truncation level ($r=8$), they are superior in accuracy to the coarsely truncated DMD solutions. Furthermore, unlike the POD-RBF solution that is trained with a first-order Euler time discretization, the NODE solutions did not exhibit any significant loss in accuracy with time, even while predicting outside the training region. It is, however, important to note that the training time for any new NODE architecture was extremely high (see Table \ref{tab:node_cylinder}) when compared to generating a POD-RBF or a DMD NIROM model, which usually required less than a minute in most cases. Such long training times may pose a significant challenge for exhaustive explorations of the design space for optimal architectures and hyperparameters, and may hinder the adoption of existing packages for automated architecture search. Thus, a concerted effort needs to be directed towards acceleration of NODE training times and towards constraining the design space by a priori identification of promising architectures.

\subsection{Shallow water equations}
The next two numerical examples involve flows governed by the depth-averaged SWE which is written in a conservative residual formulation as 
\begin{equation}\label{eq:conservative_compact}
\strongRes \equiv \pd{\vec q}{t} + \pd{\vec{p}_x}{x} + \pd{\vec{p}_y}{y} + \vec r = 0,
\end{equation}
where the state variable $\vec {q} = [h,u_x h,u_y h]^T$ consists of the flow depth, $h$, and the discharges in the $x$ and $y$ directions, given by $u_x h$ and $u_y h$, respectively. Further details about the flux vectors $\vec{p}_x$, $\vec{p}_y$ and the high-fidelity model equations are available in \cite{DFPSP2020}. The high-fidelity numerical solutions of the SWE are obtained using the 2D depth-averaged module of the Adaptive Hydraulics (AdH) finite element suite, which is a U.S. Army Corps of Engineers (USACE) high-fidelity, finite element resource for 2D and 3D dynamics \cite{Trahan2018}.

\subsubsection{Tidal flow in San Diego bay}
This numerical example involves the simulation of tide-driven flow in the San Diego Bay in California, USA.
The AdH high-fidelity model consists of $N = 6311$ nodes, uses tidal data obtained from NOAA/NOS Co-Ops website at a tailwater elevation inflow boundary and has no flow boundary conditions everywhere else. Further details are available in \cite{DFPSP2020}.

The training space is generated using $1801$ high-fidelity snapshots obtained between $t=41$ minutes to $t=50$ hours at a time interval of $\Delta t=100$ seconds. The predicted ROM solutions are computed for the same time interval with $\Delta t=50$ seconds. A latent space of dimension $265$ is generated by using a POD truncation tolerance of $\tau_{POD}= 5\times10^{-7}$ for each solution component. The RBF NIROM approximation is computed using a shape factor, $c = 0.01$. The simulation time points provided as input to the NODE model are normalized to lie in $t \in [0,1]$. The `dopri5' ODE solver is adopted for computing the hidden states both forward and backward in time. Learning from the conclusions of the cylinder example, a network consisting of a single hidden layer with $256$ neurons is deployed and the RMSProp optimizer with an initial learning rate of $0.001$, a staircase decay rate of $0.5$ every $5000$ epochs, and a momentum of $0.9$ is utilized for training the model over $20000$ epochs. For the DMD NIROM, the simulation time points are normalized to an unit time step, and a truncation level of $r=115$ is used to compute the DMD eigen-spectrum.

\begin{figure}[ht!]
\centering
 \subfloat[RBF $u_x$\label{fig:sd_rbf_u}]{%
   \includegraphics[width=0.33\columnwidth]{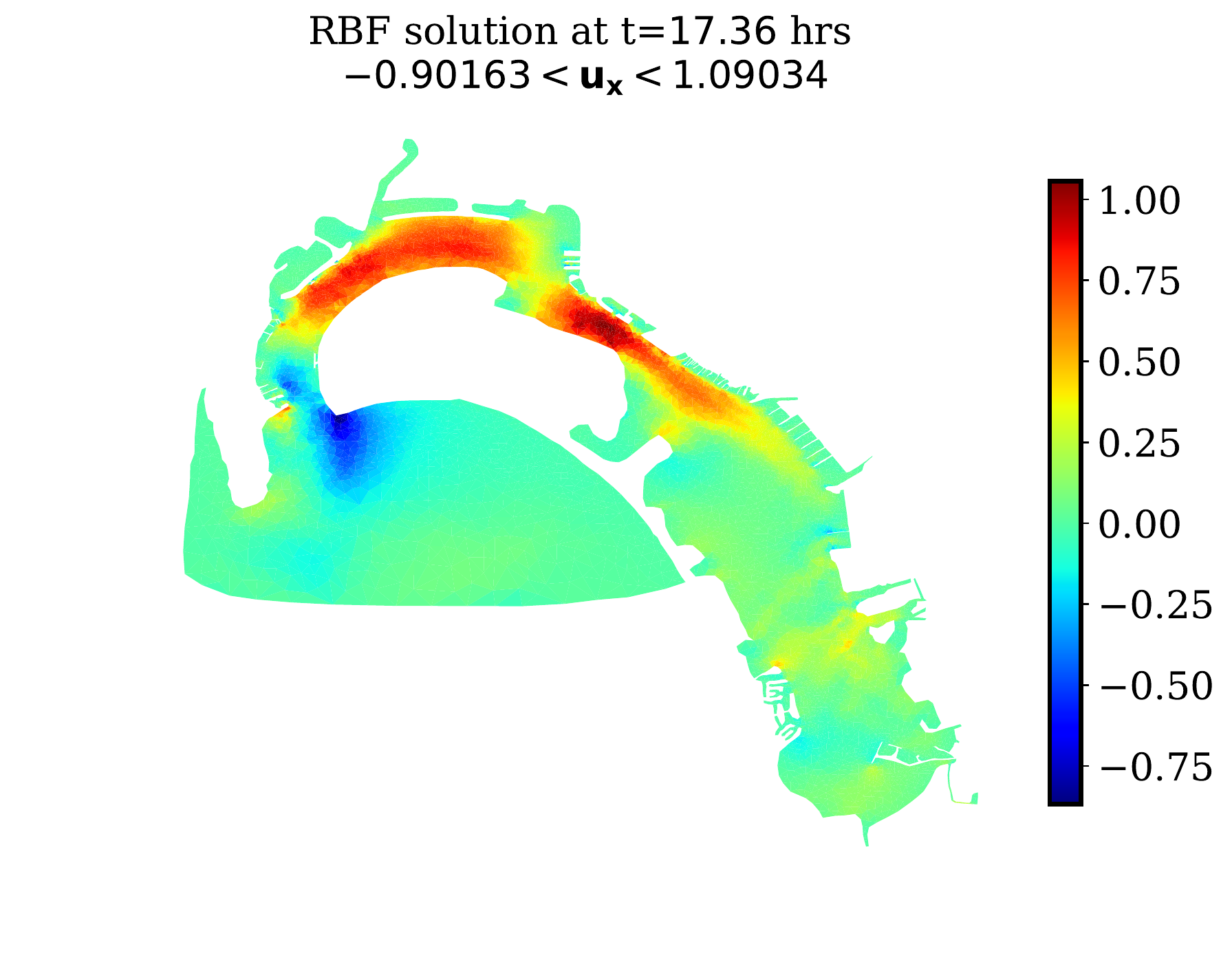}}
 \subfloat[DMD $u_x$\label{fig:sd_dmd_u}]{%
   \includegraphics[width=0.33\columnwidth]{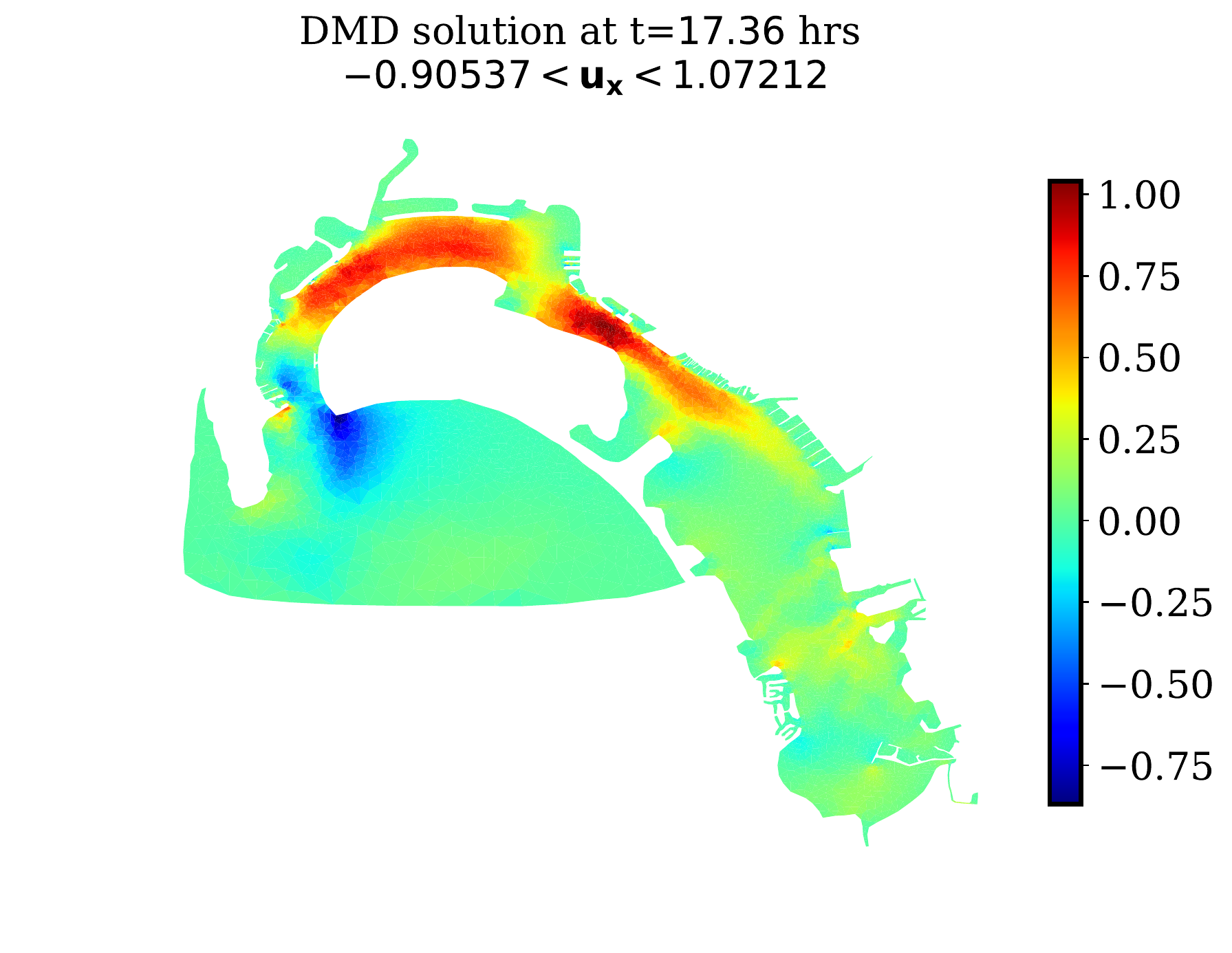}}
 \subfloat[NODE $u_x$\label{fig:sd_node_u}]{%
   \includegraphics[width=0.33\columnwidth]{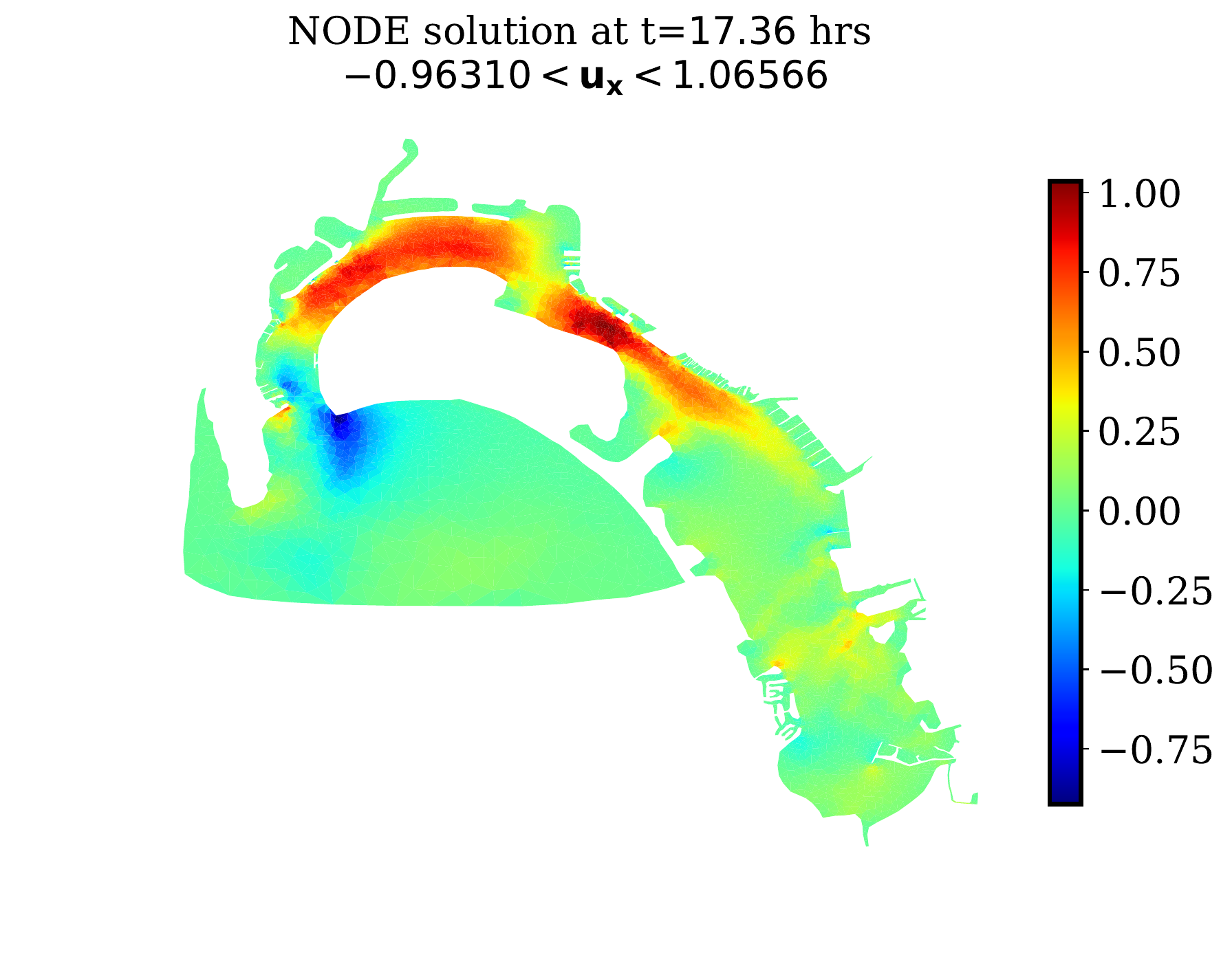}}\\
 \subfloat[RBF error\label{fig:sd_rbf_uerr}]{%
   \includegraphics[width=0.33\columnwidth]{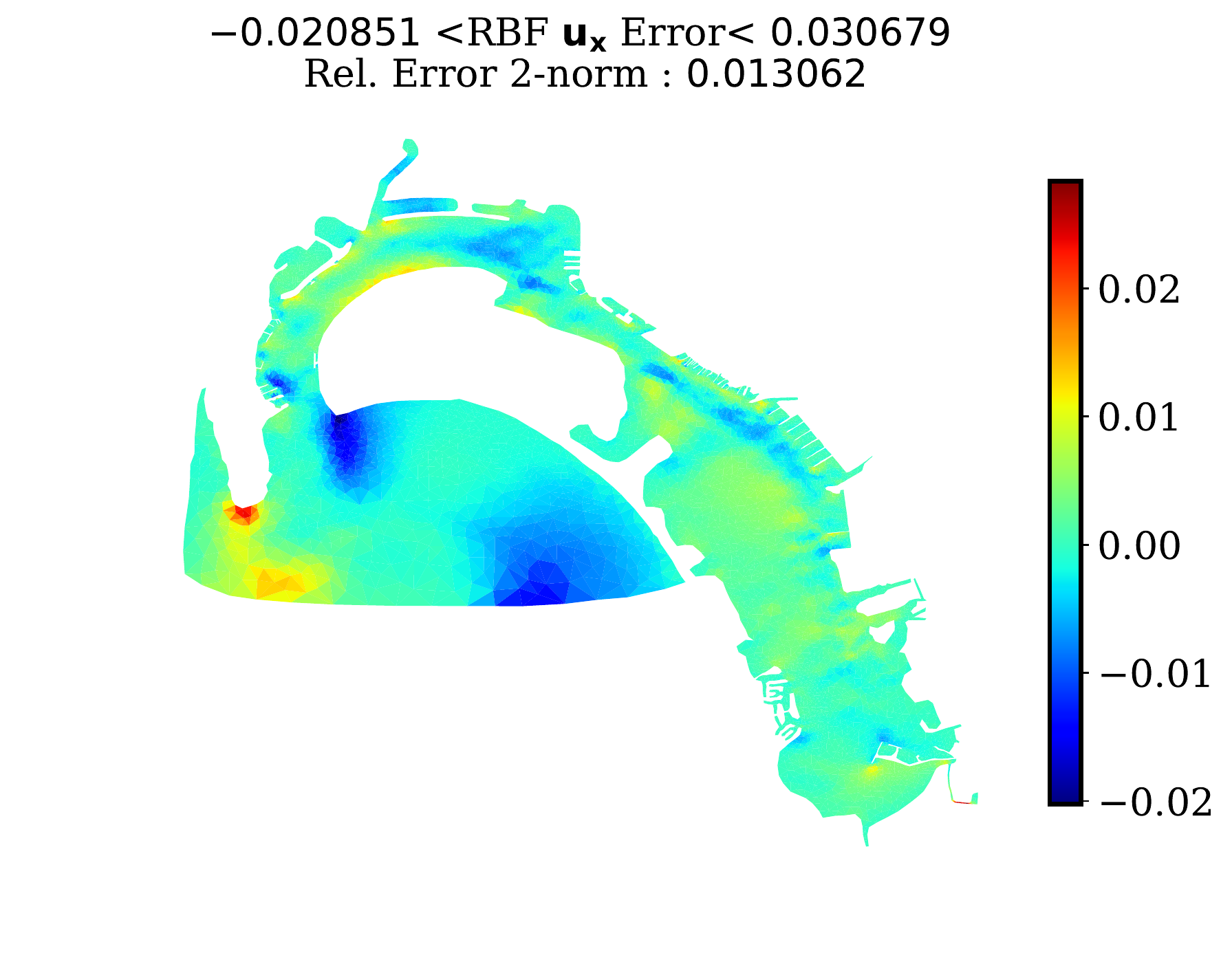}}
 \subfloat[DMD error\label{fig:sd_dmd_uerr}]{%
   \includegraphics[width=0.33\columnwidth]{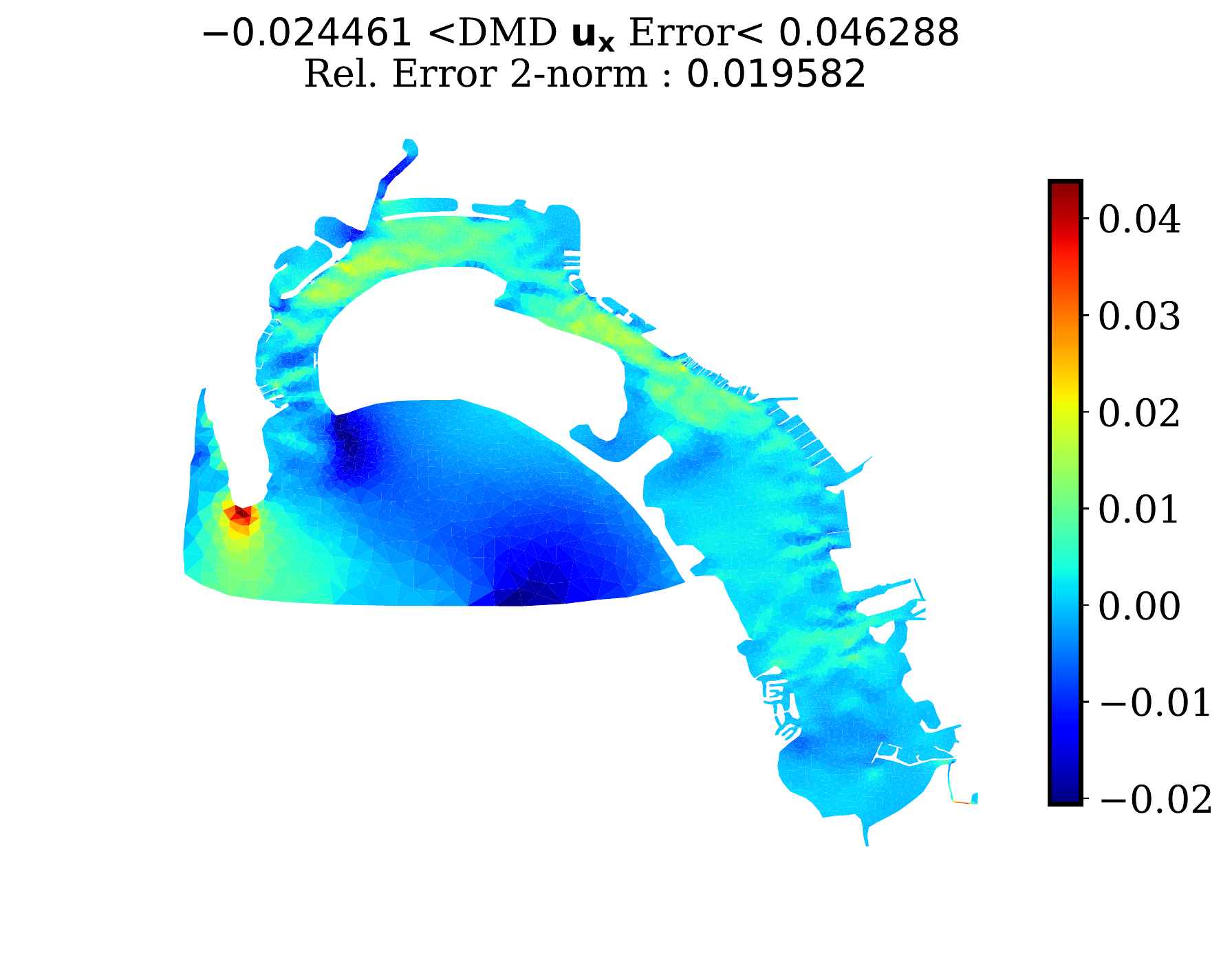}}
 \subfloat[NODE error\label{fig:sd_node_uerr}]{%
   \includegraphics[width=0.33\columnwidth]{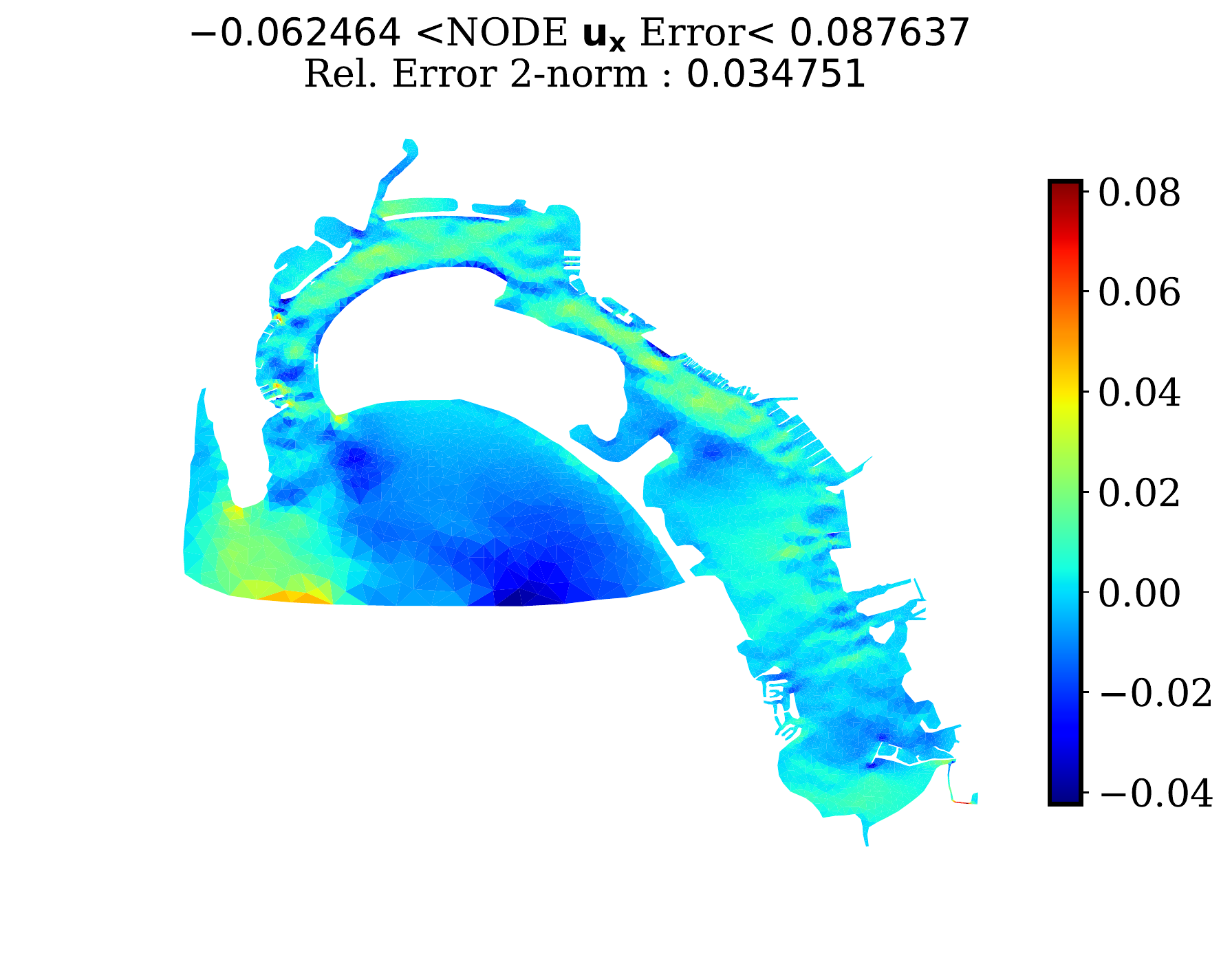}}
 \caption{NIROM solutions of $u_x$ and errors at $t=17.36$ hours for the San Diego example}\label{fig:sd_uplots}
\end{figure}

Figure \ref{fig:sd_uplots} shows the NIROM solutions (top row) for $u_x$ at $t=17.36$ hours and the corresponding error plots. 

\begin{figure}[htb]
  \begin{center}
    \includegraphics[width=0.98\columnwidth]{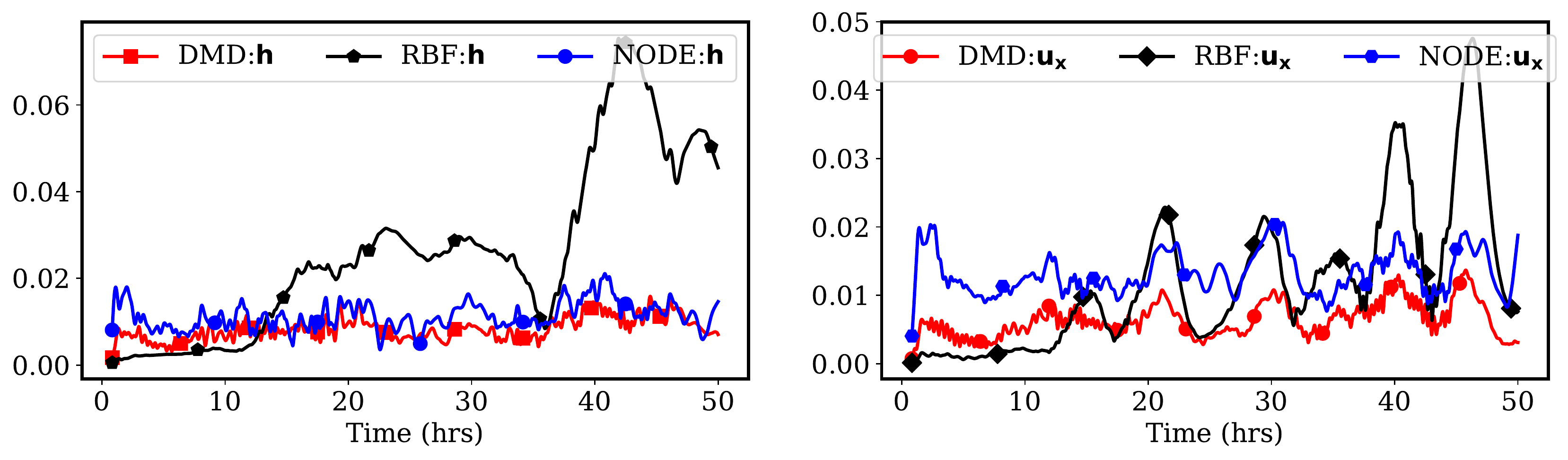}
  \end{center}
  \caption{NIROM RMSEs for the San Diego example}\label{san_diego_rms}
\end{figure}

Figure \ref{san_diego_rms} shows the spatial RMSE over time for the depth (left) and the x-velocity (right) NIROM solutions. The NODE NIROM solution has comparable accuracy to the DMD NIROM solution and unlike the RBF NIROM solution, does not exhibit any appreciable accumulation of error over time.

\subsubsection{Riverine flow in Red River}
The final numerical example involves an application of the 2D SWE to simulate riverine flow in a section of the Red River in Louisiana, USA. 
The AdH high-fidelity model uses $N = 12291$ nodes, has a natural inflow velocity condition upstream, a tailwater elevation boundary downstream, and no flow boundary along the river bank. For further details see \cite{DFPSP2020}.

The training space is generated by using $1081$ high-fidelity snapshots obtained between $t=16.67$ minutes to $t=9.3$ hours at a time interval of $\Delta t=30$ seconds. The predicted ROM solutions are computed for the same time interval with $\Delta t=10$ seconds. A latent space spanned by $54$ modes is generated by using a POD truncation tolerance of $\tau_{POD}= 0.01$ for each solution component. The RBF NIROM approximation is computed using a shape factor, $c = 0.05$. For consistency, the NODE network architecture is kept identical to the San Diego example and the training is also performed for $20000$ epochs. 
Also, similar to the previous example, the simulation time points for DMD input are normalized to an unit time step. However, a smaller truncation level of $r=30$ is used to compute the DMD eigen-spectrum.

\begin{figure}[ht!]
\centering
 \subfloat[RBF $u_x$\label{fig:red_rbf_u}]{%
   \includegraphics[width=0.33\columnwidth]{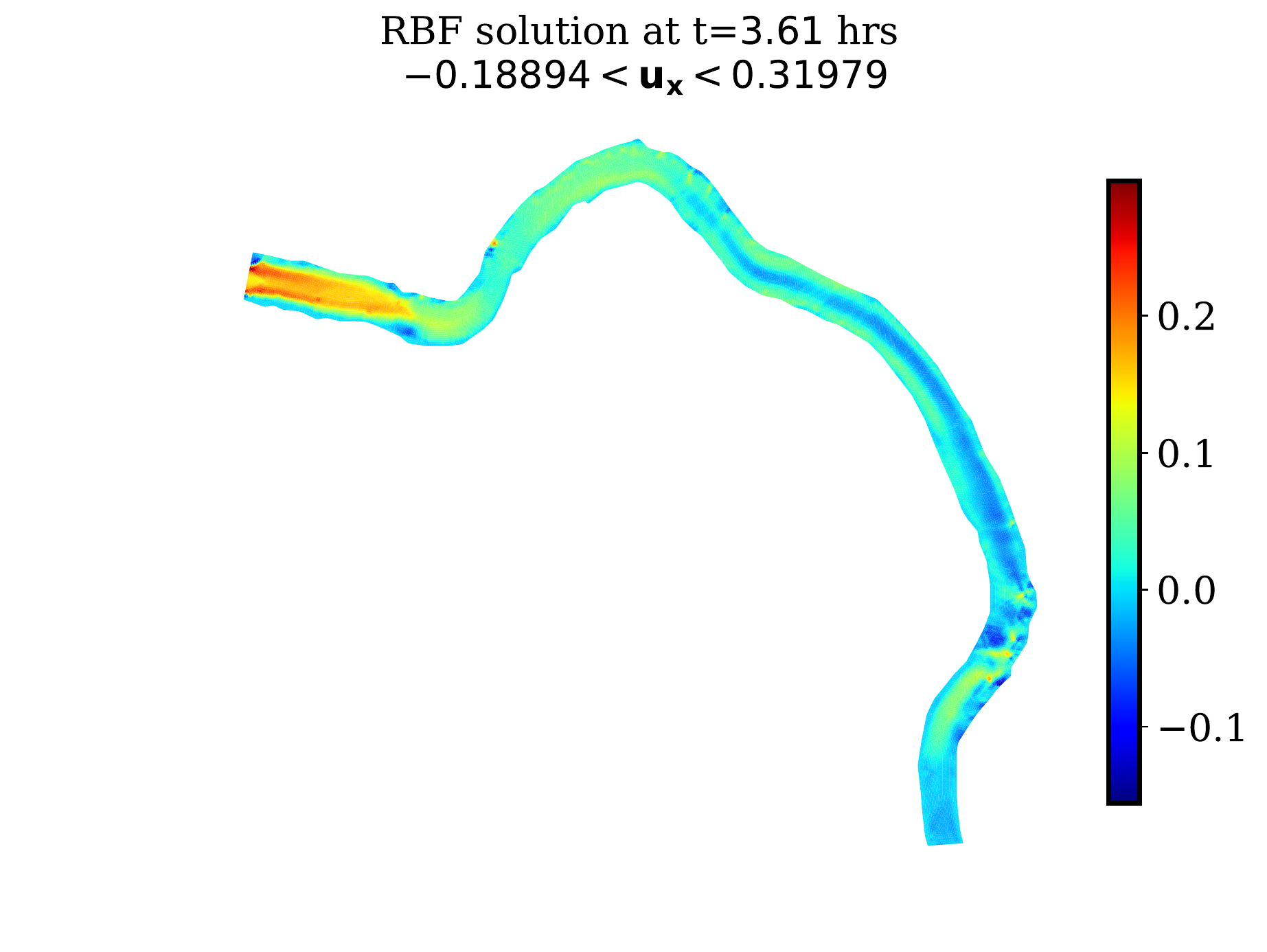}}
 \subfloat[DMD $u_x$\label{fig:red_dmd_u}]{%
   \includegraphics[width=0.33\columnwidth]{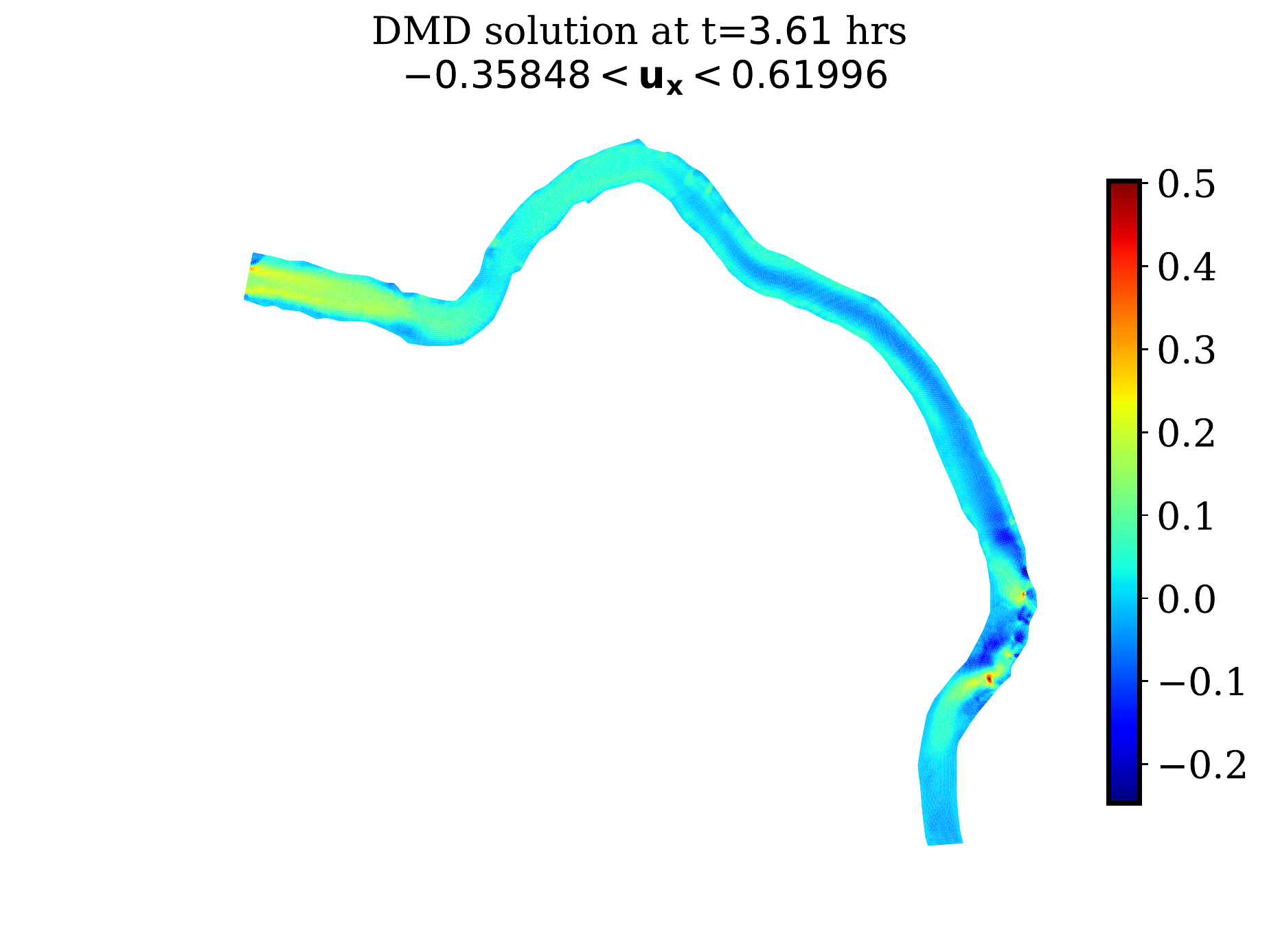}}
 \subfloat[NODE $u_x$\label{fig:red_node_u}]{%
   \includegraphics[width=0.33\columnwidth]{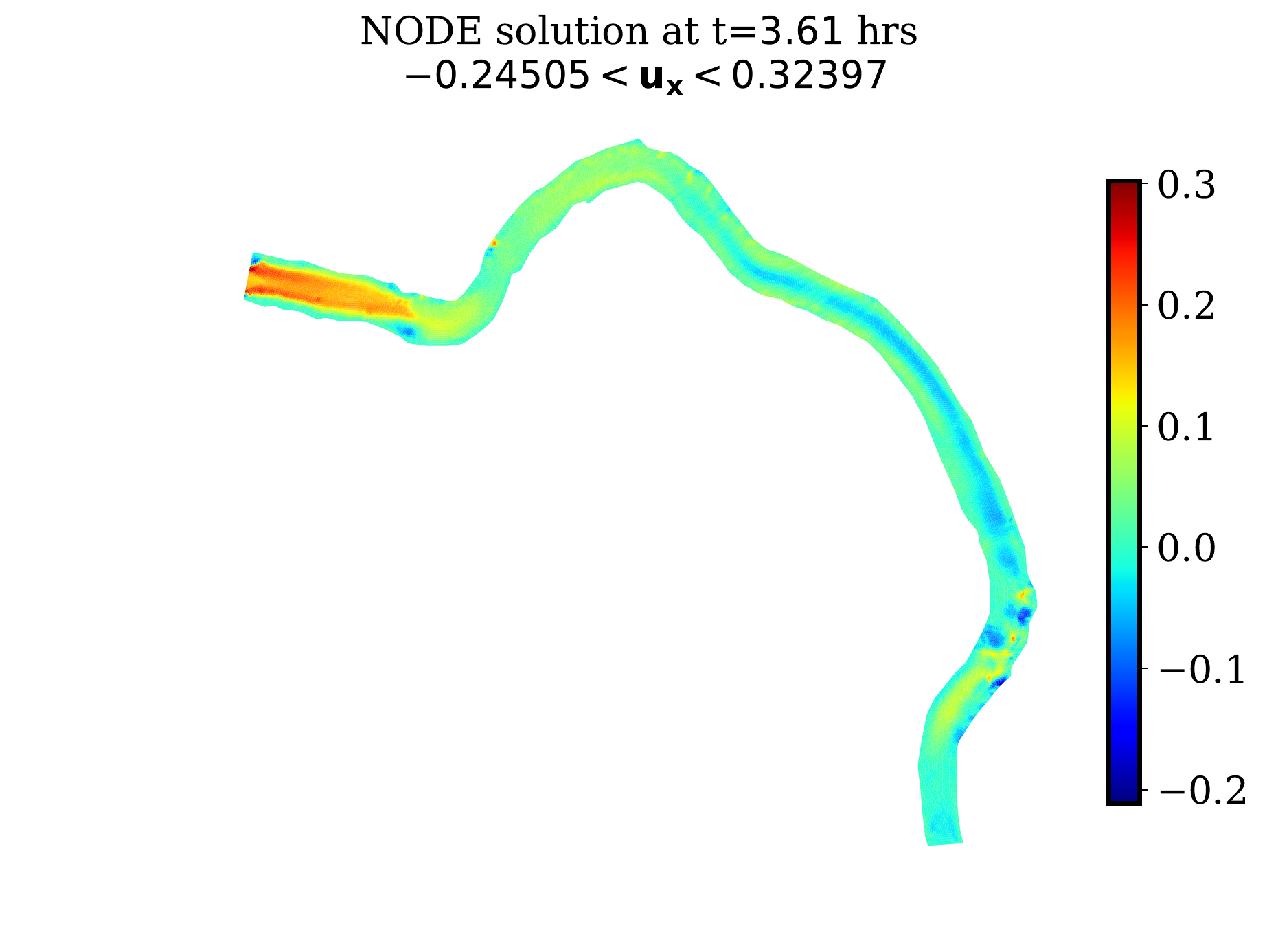}}\\
 \subfloat[RBF error\label{fig:red_rbf_uerr}]{%
   \includegraphics[width=0.33\columnwidth]{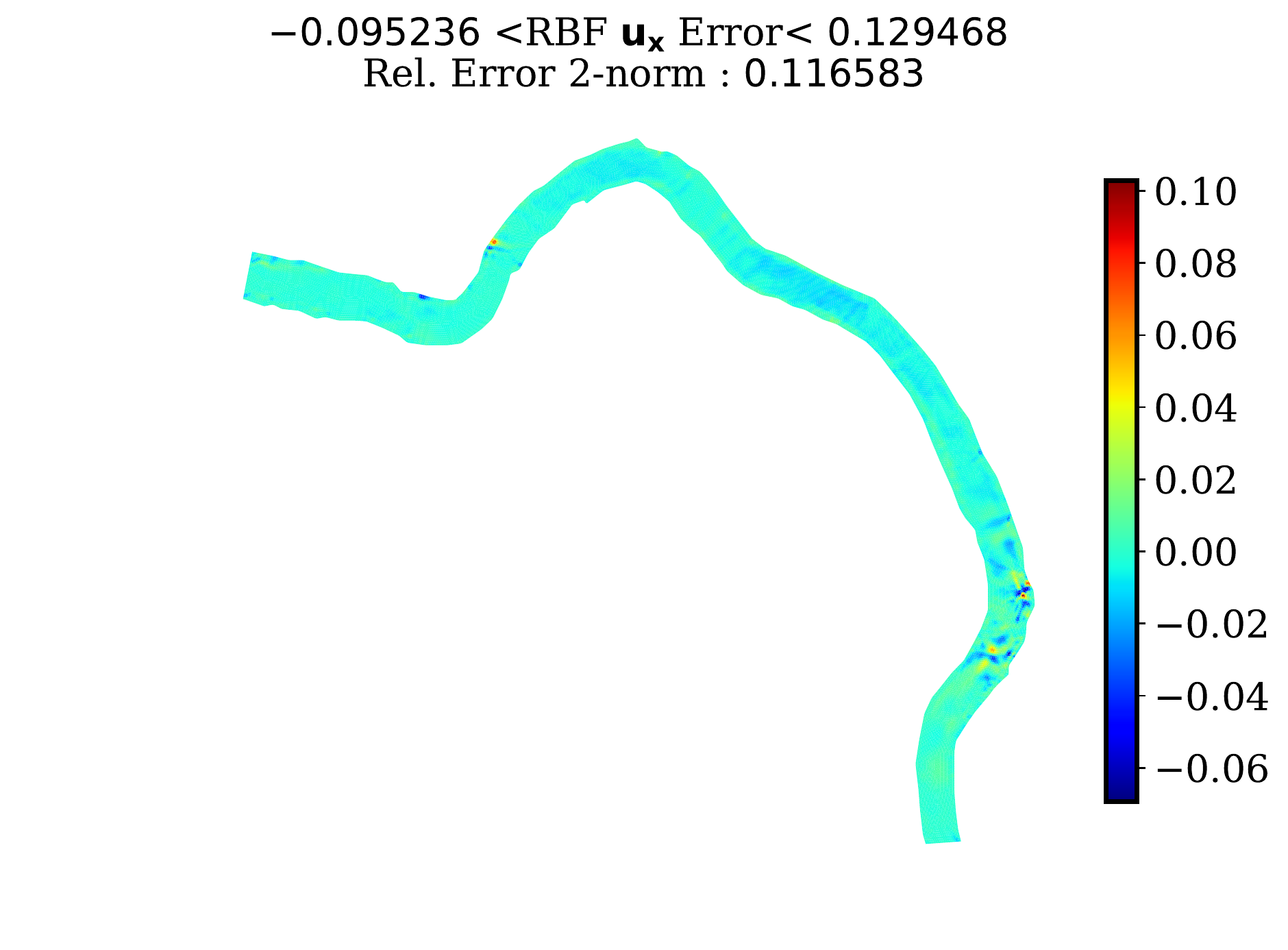}}
 \subfloat[DMD error\label{fig:red_dmd_uerr}]{%
   \includegraphics[width=0.33\columnwidth]{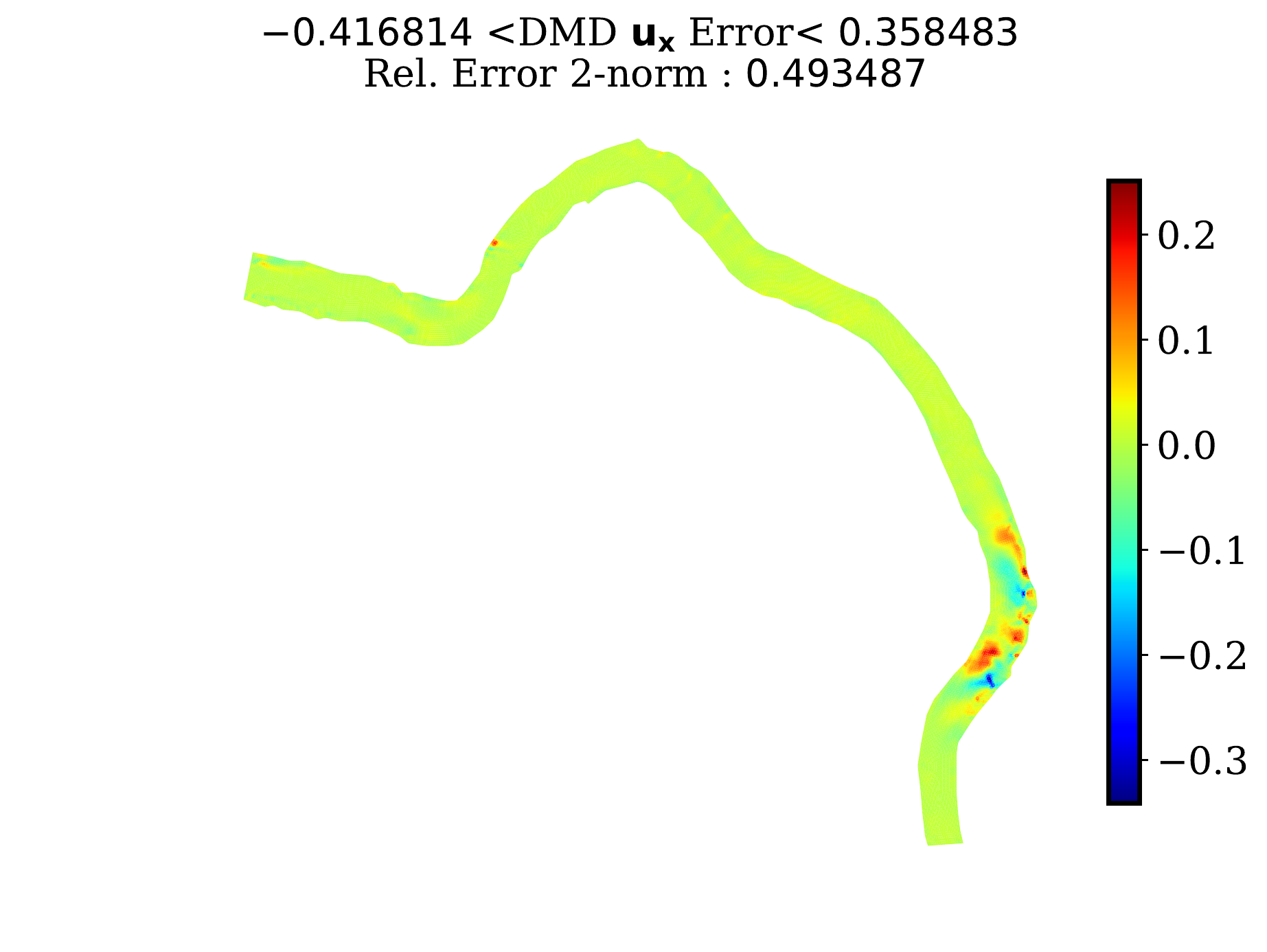}}
 \subfloat[NODE error\label{fig:red_node_uerr}]{%
   \includegraphics[width=0.33\columnwidth]{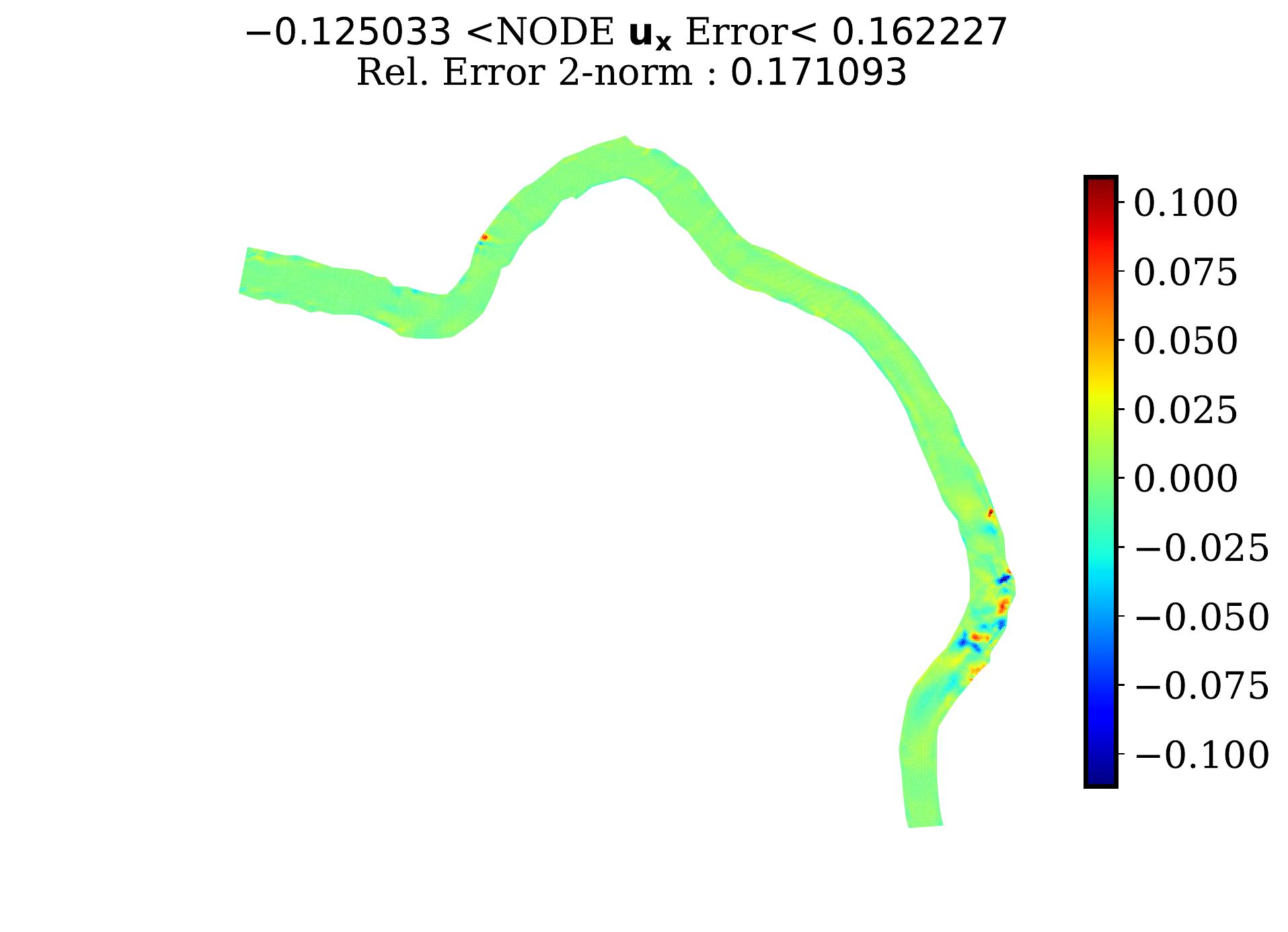}}
 \caption{NIROM solutions of $u_x$ and errors at $t=3.61$ hours for the Red River example}\label{fig:red_uplots}
\end{figure}

Figure \ref{fig:red_uplots} shows the NIROM solutions (top row) for $u_x$ at $t=3.61$ hours and the corresponding error plots. 

\begin{figure}[htb]
  \begin{center}
    \includegraphics[width=0.98\columnwidth]{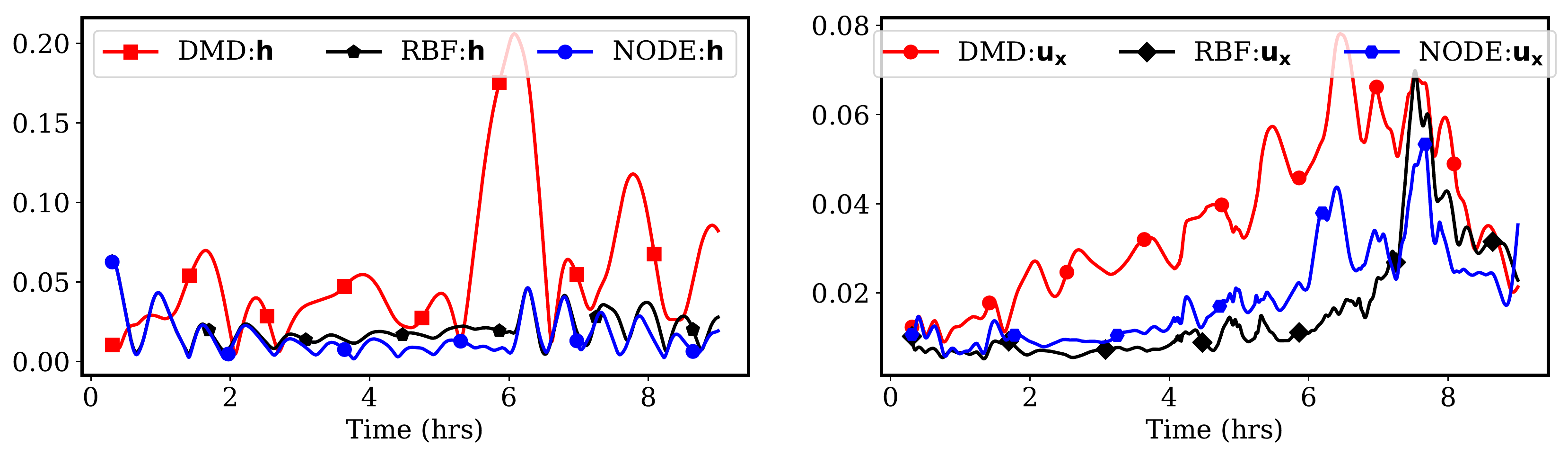}
  \end{center}
  \caption{NIROM RMSEs for the Red River example}\label{red_river_rms}
\end{figure}

Figure \ref{red_river_rms} shows the spatial RMSE over time of the depth (left) and the x-velocity (right) NIROM solutions for the Red River example. It can be seen that the DMD NIROM solution has a relatively higher RMSE owing to the lower truncation level chosen for this example, while the RBF NIROM is far more accurate. The NODE NIROM solution seems to match the performance of the RBF NIROM solution. This indicates that the NODE NIROM framework is successful with two distinct real-world flow regimes and holds promise for more widespread applicability to model the evolution of latent space dynamics.

\section{Conclusion}
We have studied Neural ODEs as a non-intrusive machine-learning algorithm to model the evolution of modal coefficients of a system of nonlinear, time-dependent PDEs in the linearly embedded latent space characterized by a truncated POD basis. Numerical experiments were carried out with a benchmark periodic flow problem governed by the incompressible Navier Stokes equations and two real-world applications of estuarine and riverine flow dynamics governed by the two-dimensional shallow water equations.The NODE formulation demonstrated a stable and accurate learning trajectory in modeling reduced basis dynamics, even in comparison to two classical ROM techniques utilizing dynamic mode decomposition and radial basis function interpolation. The DMD NIROM exhibited superior accuracy in most of the examples and was found to be most promising for long-term predictions. However, the POD-RBF NIROM technique is easily applicable to parametrized model reduction scenarios involving parametric training manifolds of very high dimension, whereas the DMD algorithm does not have a natural extension to such a setting. 
The POD-NODE formulation also produced extremely promising extrapolatory predictions for the flow around a cylinder example. This presents an exciting prospect for future exploration as even for an isolated system, unperturbed by unseen external forcings, truly extrapolative predictions of reduced order dynamics in flow regimes that do not correspond to the training data is a rare feature for most well-established ROM frameworks.

This study leads to several promising avenues of research. To begin with, an exhaustive search for an optimal NODE network architecture and optimal model hyperparameters needs to be conducted for a wide range of flow dynamics in order to gain insight of the learning trajectory and to design more generalizable NODE NIROM formulations with faster training times. With the goal of long-term predictive formulations in mind, embedding uncertainty estimates in the NODE NIROM framework might facilitate the development of adaptive models capable of re-assessing learning trajectories through in-situ measurements. The construction of a set of response functions for modeling the prediction error using machine learning \cite{Freno2019} or Gaussian Process Regression (GPR) \cite{Xiao2019} are some recent works in this direction. Another exciting field of study would be to combine the NODE framework with machine-learning strategies for the generation of nonlinear manifolds \cite{Lee2020, Kim2020} that are suitable for an efficient reduced representation of the system dynamics for advection-dominated problems and in the presence of sharp gradients where a truncated linear subspace offers a poor solution representation. All the relevant data and codes for this study will be made available in a public repository at \url{https://github.com/erdc/node_nirom} upon publication.

\section{Acknowledgements}
This research was supported in part by an appointment of the first author to the Postgraduate Research Participation Program at the U.S. Army Engineer Research and Development Center, Coastal and Hydraulics Laboratory (ERDC-CHL) administered by the Oak Ridge Institute for Science and Education through an interagency agreement between the U.S. Department of Energy and ERDC. The authors would also like to thank Dr.~Gaurav Savant for his valuable help in using the Adaptive Hydraulics suite (AdH) \cite{Trahan2018} for the high-fidelity numerical simulation of the 2D shallow water flow examples. Permission was granted by the Chief of Engineers to publish this information.

\small
\bibliography{main.bib}

\end{document}